\documentclass[a4paper,fleqn]{cas-dc}
\usepackage[numbers]{natbib}
\usepackage{multirow}
\usepackage{bbding}
\usepackage{tabularx}
\usepackage{picins}
\usepackage{algorithmic}
\usepackage{algorithm}

\def\tsc#1{\csdef{#1}{\textsc{\lowercase{#1}}\xspace}}
\tsc{WGM}
\tsc{QE}
\tsc{EP}
\tsc{PMS}
\tsc{BEC}
\tsc{DE}

\begin{document}
\begin{sloppypar}
\let\WriteBookmarks\relax
\def\floatpagepagefraction{1}
\def\textpagefraction{.001}
\let\printorcid\relax

\shorttitle{Leveraging social media news}
\shortauthors{Z. Zhang et~al.}

\title [mode = title]{BiHRNet: A Binary high-resolution network for Human Pose Estimation}

\author[1]{Zhicheng Zhang}
\cormark[1]

\ead{zczhang@bupt.edu.cn}
\cortext[cor1]{Corresponding author}
\author[1]{Xueyao Sun}
\author[1]{Yonghao Dang}
\author[1]{Jianqin Yin}

\credit{Conceptualization of this study, Methodology, Software}

\address[1]{School of Artificial Intelligence, Beijing University of Posts and Telecommunications, Beijing, China}

\begin{abstract}
Human Pose Estimation (HPE) plays a crucial role in computer vision applications. However, it is difficult to deploy state-of-the-art models on resouce-limited devices due to the high computational costs of the networks. In this work, a  binary human pose estimator named BiHRNet(Binary HRNet) is proposed, whose weights and activations are expressed as ±1. BiHRNet retains the keypoint extraction ability of HRNet, while using fewer computing resources by adapting binary neural network (BNN). In order to reduce the accuracy drop caused by network binarization, two categories of techniques are proposed in this work. For optimizing the training process for binary pose estimator, we propose a new loss function combining KL divergence loss with AWing loss, which makes the binary network obtain more comprehensive output distribution from its real-valued counterpart to reduce information loss caused by binarization. For designing more binarization-friendly structures, we propose a new information reconstruction bottleneck called IR Bottleneck to retain more information in the initial stage of the network. In addition, we also propose a multi-scale basic block called MS-Block for information retention. Our work has less computation cost with few precision drop. Experimental results demonstrate that BiHRNet achieves a PCKh of 87.9 on the MPII dataset, which outperforms all binary pose estimation networks. On the challenging of COCO dataset, the proposed method enables the binary neural network to achieve 70.8 mAP, which is better than most tested lightweight full-precision networks.

\end{abstract}

\begin{keywords}
Computer vision \sep Human pose estimation \sep Binary neural network
\end{keywords}

\maketitle

\section{Introduction}
With the rapid advancement of deep neural networks applied to various fields in computer vision, significant progress has been made in human pose estimation \cite{badrinarayanan2017segnet}\cite{lin2017feature}\cite{Sun2019DeepHR}\cite{Cheng2019HigherHRNetSR}. As models have become more and more complex, making them capable to deploy on resource-constrained edge devices is a challenging task and has become a hot issue in current research\cite{Luo2020EfficientHP}\cite{Ren2021EfficientHP}.  A variety of lightweight pose estimation networks have been designed to achieve reduction in model parameters and complexity by pruning and designing lightweight modules\cite{Yu2021LiteHRNetAL}\cite{Wang2022LitePE}. However, the weights of these lightweight models are still stored as floating-point parameters, resulting in high computation costs and large storage requirements. Binary neural network (BNN) is considered the most extreme form of quantization, as its weights and activations are represented only by ±1 \cite{Rastegari2016XNORNetIC}. The replacement of heavy floating-point multiplication and addition operations with XOR and Bitcount operations allows for drastic reduction in storage memory. Since both weights and activations are binary in BNN, it theoretically results in 58× faster convolutional operations and 32× less memory savings on CPUs than the real-valued neural networks. As a result, BNN exhibits several hardware-friendly properties, including memory savings and significant speedup \cite{Qin2020BinaryNN} .

However, most of the existing binary neural networks focus on image classification task\cite{xu2021recu}\cite{Zhuang2018StructuredBN}. The performance of binarization on different tasks varies greatly, which means that the current outstanding binary works cannot be directly applied to the human pose estimation task \cite{Qin2020BinaryNN}\cite{qin2022bifsmn}\cite{wang2020bidet}. In human pose estimation task, the model's output is a heatmap that requires pixel-level information to accurately determine the location, which is quite different from the category output of the classification task\cite{Bulat2017BinarizedCL}. The classification task aims for the category of the image and focuses on the extraction of global semantic information. While human pose estimation is a fine-grained task, which needs to determine whether the pixels on the image belong to keypoints, and involves the accurate positioning of keypoints on the human body. That means the pose estimation task requires global information to reveal the overall structure and proportional relationship of the human body, while extracting proper local information can provide more accurate joint position and posture details. Therefore, both global and local information is essential for human pose estimation. However, the effects of binarization on extracting local information for HPE is not considered in the literature. Further research is required to focus on the extraction of local information and network optimization specific to pose estimation, which is the the main motivation of this work.

To address the above challenges, we propose BiHRNet, a high-resolution human pose estimation model based on binary neural networks. BiHRNet combines recent advancements in BNN research with the HRNet structure \cite{Sun2019DeepHR} , focusing specifically on the impact of neural network binarization for fine-grained pose estimation task. To mitigate information loss resulting from binarization, knowledge distillation technique is applied, utilizing real-valued HRNet as the teacher network and the proposed binary network as the student network. The output heatmaps of the real-valued network are treated as soft labels, and the KL divergence loss is employed to guide the binary network to learn a more realistic output distribution.
This approach reduces the learning difficulty of the student network by aligning its output distribution with that of the real-valued network. Considering the limited information expression characteristics of BNN, the AWing loss\cite{Wang2019AdaptiveWL} is applied for assigning higher attention to pixels containing keypoints, which aims to prioritize the keypoint regions as important areas. In addition, we prune the model to get a more lightweight network. In terms of structural design, we introduce more binarization-friendly architectures to minimize information loss caused by binarization. We specifically redesign the bottleneck block to ensure information retention in the binary network's initial stage, where the information is reconstructed. Additionally, we develop the MS-Block, a multi-scale structure at block level, to enhance the receptive field and strengthen the information extraction capability of the networks. 

Based on these improvements, we obtain an accurary and efficient binary architecture for human pose estimation. In summary, the main contributions of this paper include:

\begin{itemize} \item We design a knowledge distillation framework to enable the pruned binary pose estimation network to learn the distribution of the output heatmaps for the real-valued network. In addition, a new loss function is proposed, which can help the binary network to obtain higher quality heatmaps. \item To reduce information loss caused by network binarization, we propose two new modules, the information reconstruction bottleneck block in the initial stage and the MS-Block retain network information in the following multiple stages. \item The proposed BiHRNet is evaluated on the MPII and COCO datasets, the experimental results show that the proposed network achieves a state-of-the-art  performance among binary pose estimation methods, which also maintain a balance between accuracy and efficiency.
\end{itemize}

\section{Related work}
Relevant prior works include studies of lightweight human pose estimation, network binarization, and knowledge distillation. 

\subsection{Lightweight Human Pose Estimation}
In the literature, the lightweight human pose estimation network design has two directions, one is to prune and quantize the high-precision pose estimation network, the other is to use the blocks of the lightweight classification networks architectures such as MobileNet\cite{Howard2017MobileNetsEC}\cite{Sandler2018MobileNetV2IR}, ShuffleNet\cite{Ma2018ShuffleNetVP}\cite{Zhang2017ShuffleNetAE} etc. For pruning and quantization high-precision network, Wang et al.\cite{Sun2019DeepHR} reduced the width and depth of the original large version of HRNet, getting a small backbone named Small HRNet. After that, Lite-HRNet\cite{Yu2021LiteHRNetAL}and Dite-HRNet\cite{Li2022DiteHRNetDL} utilized the pruned backbone of Small HRNet. Y. Wang et al.\cite{Wang2022LitePE} progressively pruned the high-precision network HigherHRNet, and verified through experiments that the high-resolution branches of the multi-branch architecture have redundancy in the network. Cutting off redundant parts of the network can reduce the amount of calculation and improve the accuracy of the network. In addition, in practical deployment scenarios, it is common to employ quantization techniques such as 16-bit quantization or 8-bit quantization to reduce the computational requirements of neural networks.

For utilizing lightweight blocks, Lite-HRNet\cite{Yu2021LiteHRNetAL} utilized a lightweight Small HRNet structure and incorporated a lightweight ShuffleBlock\cite{Zhang2017ShuffleNetAE} within the network. To further reduce the computational complexity, it designed conditional channel weighting to replace the 1×1 convolution in ShuffleBlock. BlazePose\cite{Bazarevsky2020BlazePoseOR} employed a lightweight encoder-decoder network structure to predict heatmaps for all keypoints, followed by regression to obtain the final outputs. Intel proposed an OpenPose-based network\cite{Osokin2018Realtime2M} using the design of dilated MobileNet v1\cite{Howard2017MobileNetsEC} feature extractor with depthwise separable convolutions and a lightweight refinement stage with residual connections.  

The above works focus on full-precision models that are lightweight in terms of network structure. However, these models still utilize 32-bit floating point numbers for multiplication and addition operations, which can be computationally expensive. In contrast, our proposed approach in this study introduces a binarized pose estimation network that significantly reduces the computational resource requirements. 

\subsection{Network Binarization}
Among the existing network compression methods, quantization represents the weights with low precision, which is a promising technique that yields highly compact models. Network binarization is considered as the most extreme quantization, for its weights and activations are quantized to 1 bit\cite{Rastegari2016XNORNetIC}. For this reason, compared to full-precision networks, BNNs have limited expressive ability. In addition, since the Sign function is not differentiable at 0, and the derivative of the function at non-zero points takes the value 0 everywhere, the gradient transfer using traditional differentiation methods becomes problematic. For this problem, many methods have been used to alleviate the impact of network binarization from different perspectives. M. Courbariaux et al.\cite{Courbariaux2016BinaryNetTD} designed straight-through estimation (STE) to enable BNN to learn gradient through backward propagation. Many subsequent works improved and optimized STE, designing STE variants to approximate the gradient of the Sign function.  Z. Liu et al.\cite{Liu2018BiRealNE} designed a continuous piecewise function ApproxSign to replace the Clip function, which was closer to the Sign function than the Clip function, thus reducing the information loss in directional propagation. H. Qin et al.\cite{Qin2019ForwardAB} proposed Error Decay Estimator(EDE), which gradually approximated the Sign function at different training stages, and used EDE to replace the Sign function for back propagation, making the entire training process smoother. The method proposed by K. Helwegen et al.\cite{Helwegen2019LatentWD} discussed the functional role of latent weights in BNNs and proposed a specialized optimizer BOP to transform binary states. Bi-real Net\cite{Liu2018BiRealNE} added additional shortcuts to reduce the information loss caused by binarization. J. Bethge et.al\cite{Bethge2020MeliusNetCB} designed two binarization-friendly modules to enhance the quality and capacity of the network. B. Martinez et al.\cite{Martnez2020TrainingBN}used real-valued activation before binarization to calculate the scaling factor, and the factor was multiplied by the output of the binary convolution to improve the representation ability of the BNN.

For human pose estimation task, only few works have attempted to apply binary neural networks for the network construction. Bulat A et al.\cite{Bulat2017BinarizedCL} studied the effect of neural network binarization on pose estimation for the first time, and proposed a hierarchical and multi-scale residual architecture, which had parallel paths with receptive fields of different sizes. Bulat A et al. \cite{Bulat2019MatrixAT} made use of matrix decomposition to binarize the weight tensor of each layer, and the method was evaluated on MPII. Y. He et al. \cite{He2022BinarizingBC} treated network binarization as a binary classification problem and used a multi-layer perceptron (MLP) as the classifier.  However, these works did not specifically consider the impact of network binarization on fine-grained tasks. During the binarization process, too much local information required for keypoint localization is lost. In our work, we design more binarization-friendly modules to ensure the retention and transmission of information.

\subsection{Knowledge Distillation}
Knowledge distillation (KD)\cite{hinton2015distilling} is a technique that can train a student network to learn the performance of the teacher network. Compared with purely supervised training, it can provide a more comprehensive training signal, which works well for training small networks.

In binary network training, knowledge distillation is often used to bridge the output distribution gap between full precision network and binary network. Z. Liu et al.\cite{Liu2020ReActNetTP} used the outputs of the full-precision network as soft labels to assist the training of the binary neural network. N. Guo et.al \cite{Guo2022JoinTH} proposed a new knowledge distillation method that alleviated the overfitting problem when training binary neural network models with high accuracy; while Real2BinaryNet\cite{Martnez2020TrainingBN} trained the network using three stages to distill binary network asymptotically. In 2D human pose estimation networks, F. Zhang et al.\cite{Zhang2018FastHP} firstly introduced knowledge distillation to train a lightweight Hourglass network. Z. Li et al.\cite{Li2021OnlineKD}  established an online knowledge distillation framework that distilled information in a one-stage manner. In this work, we propose a real-to-binary distillation framework for human pose estimation.

\section{BiHRNet}
\vspace{\baselineskip}
\subsection{Preliminaries}
BNN uses ±1 to represent the weights and activations instead of using 32-bit floating point numbers, and usually directly uses the Sign function to binarize the weights and activations of the network\cite{Rastegari2016XNORNetIC}:

\begin {equation}
\label{sign1}
{Sign(x)} = 
\begin {cases}
-1 &{x < 0} \\
{1}&{x \geq 0}
\end {cases}
\end{equation}

Matrix multiplication is the core operation in convolution. In BNN, the calculation process of convolution operation can be expressed by the following formula:

\begin{equation}
\label{conv1}
z=\alpha Q_{b}(W)^{T} Q_{b}(A)=\frac{\|W\|_{L 1}}{k \times k \times c_{i n}} \operatorname{Bitcount}\left(W_{b}^{T} \oplus A_{b}\right)
\end{equation}

\noindent where \emph{W} and \emph{A} represent real-valued weights and activations respectively; $Q_b$ represents Sign function, through which the real-valued weights and activations are transformed into binary values; $\alpha$ is the calculated proportional factor, for reducing the quantization error caused by weight binarization; $k$ refers to the size of the convolution kernel, and $c_{in}$ is the number of input channels; $\oplus$ means XOR operation. The resulting binary activations and weights are both one-bit, so logical operations (XOR and Bitcount operations) can be used instead of ordinary floating-point operations, which speeds up the inference process.

Pose estimation is a position-sensitive task. HRNet maintains a high resolution from beginning to end, which has multiple parallel branches of different resolutions that continuously exchange information. The structure collects semantic information and accurate position information at the same time. In this work, HRNet is choosen as the backbone of the network due to its powerful feature extraction ability. A set of general principles for training binary neural networks is applied\cite{Rastegari2016XNORNetIC}\cite{Liu2018BiRealNE}\cite{Liu2020ReActNetTP}, where the the ApproxSign function is utilized to binarize the full-precision network\cite{Liu2018BiRealNE}. 

The performance of the real-valued HRNet and the binarized network on the MPII dataset is shown in Table 1. It can be observed that the binarization of the network suffers a significant performance drop in all evaluation metrics, compared with its real-valued counterpart. The reduction in accuracy is due to the characteristic that the binary neural network only uses ±1 for weights and activations. As a result, its information extraction and expression capabilities are greatly impaired.  In order to promote the estimation accuracy and reduce the information loss caused by network binarization, several techniques are developed to enhance the BiHRNet.

\begin{figure*}
	\centering
        \includegraphics[width=\textwidth]{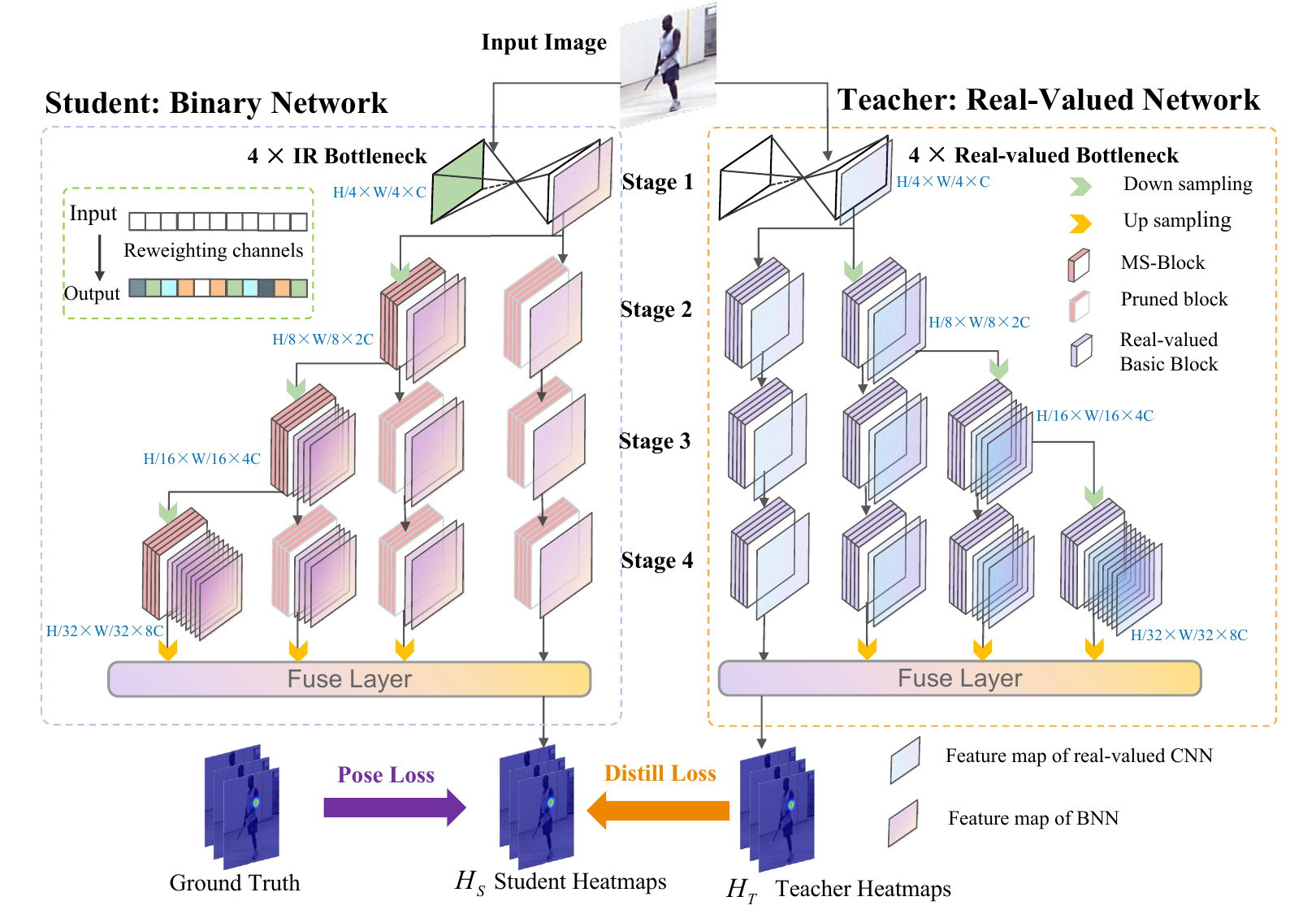}
	\caption{Overall structure of BiHRNet. The real-valued teacher network supervises the binary student network using its informative output heatmaps. The building blocks of the binary student network include IR Bottleneck and MS-Block.}
	\label{FIG:1}
\end{figure*}

\begin{table}[width=.9\linewidth,cols=4,pos=h]
\caption{Performance of real-valued and binary HRNet on MPII val. The input resolution is 256×256.The structure is HRNet-w32, neither of them use pre-trained model.}
\label{tbl1}
\begin{tabular*}{\tblwidth}{@{} LLLL@{} }
\toprule
\textbf {Crit.}  & \textbf{PCKh}   & \textbf{PCKh\protect\@0.1} & \textbf{\protect\#Par} \\
\midrule
HRNet(Real)   & 87.734 & 33.685   & 28.5M \\
HRNet(Binary) & 76.432 & 19.761   & 28.5M \\
\bottomrule
\end{tabular*}
\end{table}

\subsection{Knowledge Distillation for Binary HPE Network Training}
As shown in Figure 1, BiHRNet is a 4-stage network consisting of a high-resolution main branch with the highest resolution and three subquent branches with high-to-low resolution, which are added one by one in parallel at the beginning of each new stage to the network. Each newly added branch has half the resolution and double channels compared with the previously added branches. Among the four stages of BiHRNet, the first stage contains two 3×3 convolutions and four Information Reconstruction Bottleneck. The following three stages consist of a series of cross-resolution modules, which are composed of Multi-Scale Basicblocks and multi-stage fusion layers that exchange information across all branches. The branch maintaining the highest resolution representation provides the final output of the backbone for the subsequent pose estimation. It has been shown that the high-resolution branches are redundant for models in the low-computation region\cite{Wang2022LitePE}. We conduct network pruning to reduce redundancy and enhance the model's efficiency. As presented in Fig 1, in our multi-branch architecture, denoted as $\left\{B_1,B_2,B_3,B_4\right\}$, each $B_n=[b_1, b_2,...,b_n]$ represents the number of blocks at stage $n$, with $b_i$ indicating the number of blocks in branch $i$.  Initially, the original backbone configuration is $\left\{[4],[4,4],[4,4,4],[4,4,4,4]\right\}$. After pruning, the network configuration becomes $\left\{ [4],[0,4],[0,0,4],[0,0,0,4] \right\}$. This pruning strategy reduce the number of model parameters and improve computational efficiency, and has been verified in subsequent experimental results.

Since the information of feature map in BNNs is not comprehensive enough, the ability for BNNs to carry information through the network structure is weak. We use knowledge distillation to make BNNs for HPE obtain more comprehensive distribution information. We use the real-valued network with stronger information retention ability as the teacher, whose heatmap output is used as the soft label, and transfer both the distribution information and location information to the binary student network. In the training phase, the teacher network passes the distribution information to the student network. While in the inference phase, only the binary student network is used. The loss function used in training is composed as follows:\\

\noindent \textbf{Pose loss Function\\}
MSE loss function is  commonly used in pose estimation\cite{Sun2019DeepHR}\cite{Xiao2018SimpleBF}\cite{Zhang2019SimpleAL}, where the attension for all the pixels is considered as the same. However, as the background part in a heatmap is much larger than the part occupied by the Gaussian map of the predicted keypoints, the Awing loss function is utilized in BiHRNet to obtain the gap between ground truth and predicted heatmaps, which can significantly improve the quality of heatmap regression results\cite{Wang2019AdaptiveWL}. The loss formula of the Awing loss is as follows:\\

\begin {equation}
\label{Awing}
{L_{Awing}(y,\hat{y})} = 
\begin {cases}
\omega {ln}(1+ {\left| \frac{y - \hat{y}}{\varepsilon} \right|}^{\alpha -y} &{if \left| {y - \hat{y}} \right| < \theta} \\
{A \left| {y - \hat{y}} \right| - C}&{otherwise}
\end {cases}
\end{equation}

\noindent among them, $y$ and $\hat{y}$ are the pixel values on ground truth heatmap and predicted heatmap respectively; $\omega$ , $\varepsilon$ , $\alpha$ and $\theta$ are all positive values; $A$ and $C$ are used to smooth the loss function at ${\left| y - \hat{y} \right|} = \theta$ , where they are formulated as: 
\begin {equation}
\label{A}
A=\omega(1 /(1+(\theta / \varepsilon)^{(\alpha-y)}))(\alpha-y)((\theta / \varepsilon)^{(\alpha-y-1)})(1 / \varepsilon)
\end{equation}

\begin{equation}
\label{C}
C=\left(\theta A-\omega \ln \left(1+(\theta / \varepsilon)^{(\alpha-y)}\right)\right)   
\end{equation}
\noindent $\theta$  is used as a threshold to switch between linear and nonlinear phases. When ${\left| y - \hat{y} \right|} < \theta$ , it is considered that the gap between the output and the ground truth is too small and a more powerful influence is needed.\\

 \noindent \textbf{KL Loss\\}
 The pixel value in the heatmap represents the probability that the pixel falls on the keypoint. We use the pixel-level KL (Kullback-Leibler) divergence loss to minimize the distribution difference between the output heatmap of the real-valued teacher model and the binary student model. The binary neural network is expected to obtain a similar output distribution from a real-valued network:

 \begin{equation}
\begin{split}
\label{KLLoss}
L_{K L}&=\frac{1}{n} \sum_{i \in M} \sum_{j=1}^{n} K L\left(p_{r}^{i}\left(X_{j}\right), p_{b}^{i}\left(X_{j}\right)\right) 
\\ &=\frac{1}{n} \sum_{i \in M} \sum_{t=0}^{T} p_{r}^{i}\left(X_{j}\right) \log \left(\frac{p_{r}^{i}\left(X_{j}\right)}{p_{b}^{i}\left(X_{j}\right)}\right)
\end{split}
\end{equation}

\noindent where $p_{r}^{i}(X_{j})$ and $p_{b}^{i}(X_{j})$ are defined as the probability of the i-th pixel in the heatmap generated by the real-valued teacher model and the binary student model, $n$ is the number of batches, and $M$ represents all pixels in the heatmap. The $L_{KL}$ loss is defined as the KL divergence between $p_{r}^{i}(X_{j})$ and $p_{b}^{i}(X_{j})$ .\\

\noindent \textbf{Overall loss function\\}
We formulate the overall loss function during training process as:

\begin{equation}
\label{OverallLoss}
L_{Total} = {\alpha} L_{Awing} + (1- {\alpha}) L_{KL}
\end{equation}

\noindent where  $\alpha$ is the balancing weight between the two loss terms. Through the design of the overall loss function, the binarized student network obtains the labeled real-value information through $L_{AWing}$ , and obtains the output distribution of the real-valued network through $L_{KL}$ .
The training process of the network is summarized in Algorithm 1.

\begin{algorithm}[H]
\caption{Binary Network Knowledge Distillation}\label{alg:alg1}
\begin{algorithmic}
\STATE 
\STATE {\textbf{Input:}} Labeled training dataset \textbf{$D$} ; training rounds \textbf{$\varepsilon$} ; 
\STATE \hspace{1.0cm} {teacher network \textbf{$N_T$}; student network \textbf{$N_S$}  .}
\STATE {\textbf{Output:}} Binary student network output heatmap.
\STATE {\textbf{Initialize:}} Epoch=0; $N_S$ Initialization.
\STATE \hspace{0.5cm}$ \textbf{While e \textless $\varepsilon$ } $
\STATE \hspace{1.0cm}$ Compute\ the\ output\ heatmap\ of\ \textbf{$N_T$}; $
\STATE \hspace{1.0cm}$ Compute\ the\ output\ heatmap\ of\ \textbf{$N_S$}; $
\STATE \hspace{1.0cm}$ Compute\ loss\ according\ to\ (3),\ (6),\ (7);$ \
\STATE \hspace{1.0cm}$ Update\ the\ model\ parameters\ of\ \textbf{$N_S$} ;$ \
\STATE \hspace{1.0cm}$ e = e + 1 $
\STATE \hspace{0.5cm}$ \textbf{End while} $
\STATE {\textbf{Model Deployment:}} Use binary $N_S$ pose estimator.
\end{algorithmic}
\label{alg1}
\end{algorithm}

\subsection{Block design}
\vspace{\baselineskip}
\subsubsection{Basic binary convolution}
The smallest module of the network is convolution block, which forms Bottleneck and Basicblock. The activations of BNN need to be binarized before convolution, so it is important to use specific structure to ensure the information retention, where the binary convolution module has its own widely used settings\cite{Liu2020ReActNetTP}. The commonly used structure is shown in Figure 2 (b), where we call it Binary Unit to distinguish it from the direct binarized convolution. The use of residual connection, batch normalization and PReLU in it can better retain information compared to original binary convolution shown in Figure 2 (a).

\begin{figure}[ht]
\centering
\includegraphics[width=3.5in]{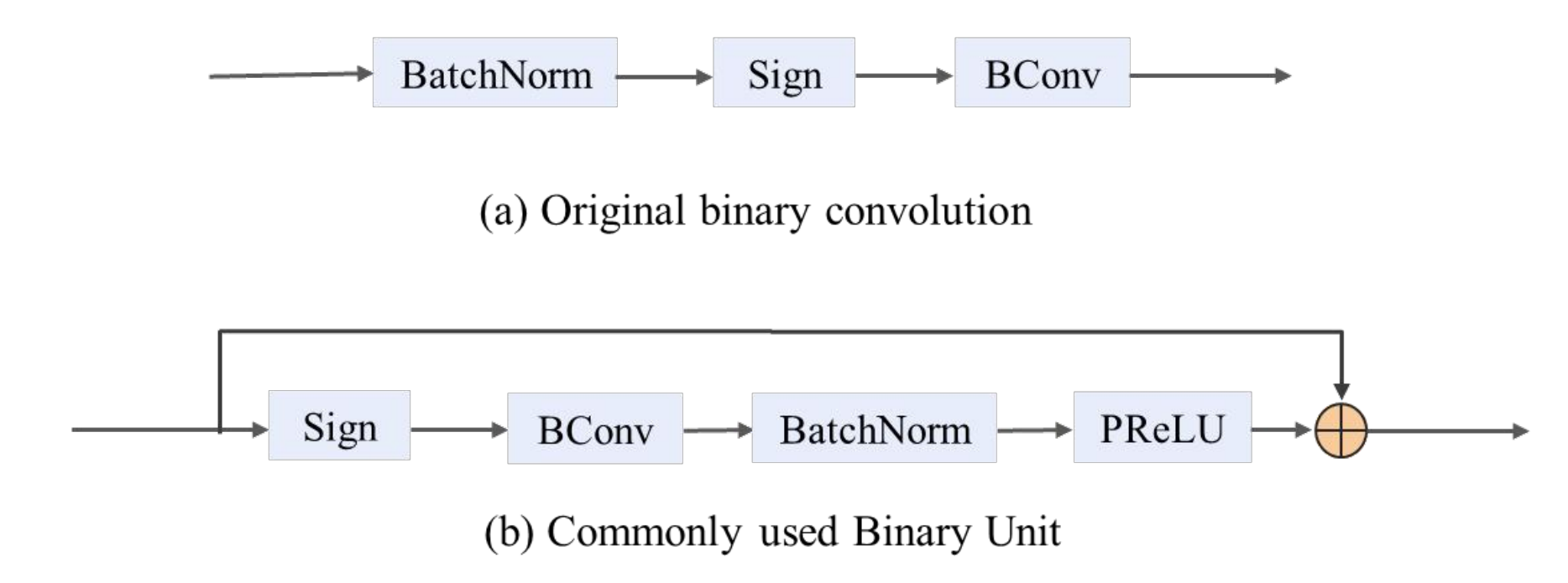}
\centering
\caption{\centering Binary convolution module}
\label{FIG:2}
\end{figure}

Basicblock and Bottleneck are the building blocks that compose the four stages of HRNet. Direct binarizing these building blocks will cause large information loss due to lack of information interaction and channel reduction. Therefore,  it is necessary to consider designing more efficient and binarization-friendly basic modules to form a binary network, which aims to reduce the loss of information passed between network layers.

\vspace{\baselineskip}
\subsubsection{Binary Bottleneck for Information Reconstruction}
The Bottleneck constitutes the first stage of the network, which contains three convolutional layers of 1×1, 3×3 and 1×1, as shown in Figure 3 (a). The number of channels decreases first and then increases, which is designed to reduce the number of parameters. Since the Bottleneck block constitutes the initial stage of the network and is closest to the input, it is necessary to ensure the information loss is within a controllable range. This ensures a smooth training process for the subsequent three stages of the network. 
In order to reduce the loss of Bottleneck information, we adopt two methods to achieve a binary Bottleneck structure with less information loss. Instead of directly binarizing the network, we replace the convolutional block with the more binarization-friendly Binary Unit mentioned above. 

For reducing information loss, it has been proved that the closer the output distribution of binary convolution is to its real-valued counterpart, the less information loss will be caused by binarization\cite{Liu2020ReActNetTP}. The use of Binary Unit can reduce part of the information loss, while the representation ability of binary network is still limited. In order to make the feature distribution in the initial stage of the binary network close to the real valued counterpart, we add a channel attention module SE Block\cite{Iandola2016SqueezeNetAA} to each convolution Unit. SE Block contains two fully connected layers. For relatively shallow classification network, such as ResNet-18\cite{He2015DeepRL}, the amount of additional calculation and parameter is small. However, pose estimation task needs a larger network scale to learn more complex feathers and pixel level information to get precise heatmap. For example, ResNet-50 is used in SimpleBaseline\cite{Xiao2018SimpleBF}.  If SE Block is added to each convolution, the amount of parameters and calculation cost will increase a lot, which cannot be ignored. Therefore, we hope attention can be added to the block level not convolution level, to make the feathers in the first stage closer to features in real-valued networks.

Specifically, the proposed Bottleneck for information reconstruction we designed is shown in Figure 3 (b). In addition to the preserved residual connection, we introduce the SE Block to process the input feature of the block. This allows the attention block to capture information changes from the input data and dynamically adjust the output distribution of the Bottleneck block through learning. The SE Block enhances the adaptability of the model by selectively emphasizing or suppressing features in the input, based on their relevance to the task at hand. This helps to improve the overall performance and effectiveness of the model in capturing and utilizing relevant information. The structure achieves superior performance by computing the scale factor for each channel in a data-driven manner. The channel weight can be expressed as: 

\begin {equation}
\label{s}
s=\operatorname{Sigmod} (FC(\operatorname{Re}LU( FC(\text { Pooling }(x)))))
\end{equation}

\noindent with the attention weight $s$ , the model pays more attention to channels with more information in the initial stage.

\begin{figure}[ht]
\centering
\includegraphics[width=0.5\textwidth]{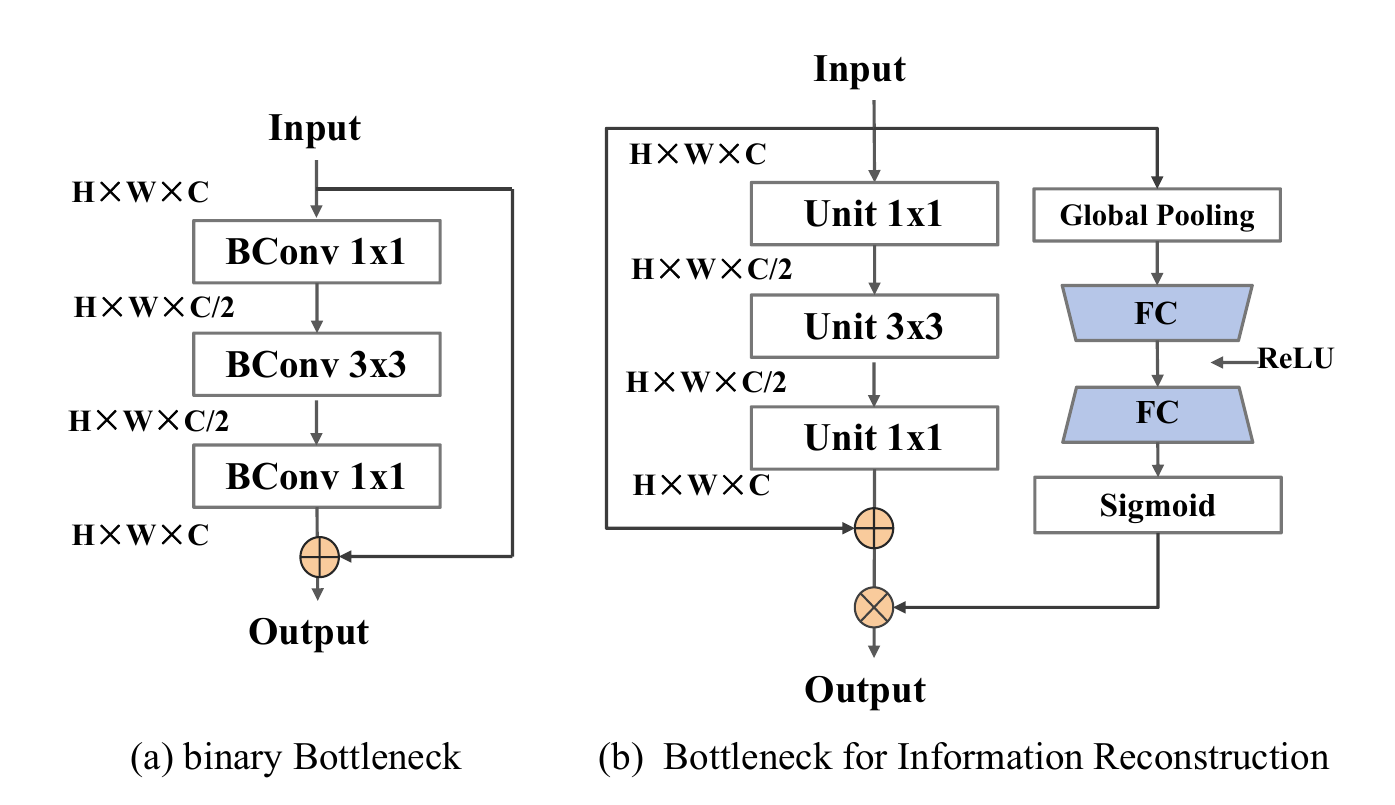}
\centering
\caption{\centering Bottleneck module design}
\label{FIG:3}
\end{figure}

\vspace{\baselineskip}
\subsubsection{Multi-scale Basicblock: MS-Block}
CNN extracts features through layer-by-layer convolution. In this process, an important concept is the receptive field, where more attention should by paid in binary network design. If the receptive field is too small, only local features can be observed, which is magnified in binary networks for its weak information retention ability. HRNet is a parallel multi-branch network structure, the information extracted from different stages mixes together in fuse layer between adjacent stages. The basic module that constitutes the 2nd, 3rd and 4th stages of HRNet is the Basicblock, which consists of two 3×3 convolutions, as shown in Figure 4 (a). Directly binarizing it leads to small receptive field and limited information extraction. In BiHRNet, we construct a basic module with a multi-scale convolution block with stronger information extraction ability, as shown in Figure 4 (b).

\begin{figure}[ht]
\centering
\includegraphics[width=0.5\textwidth]{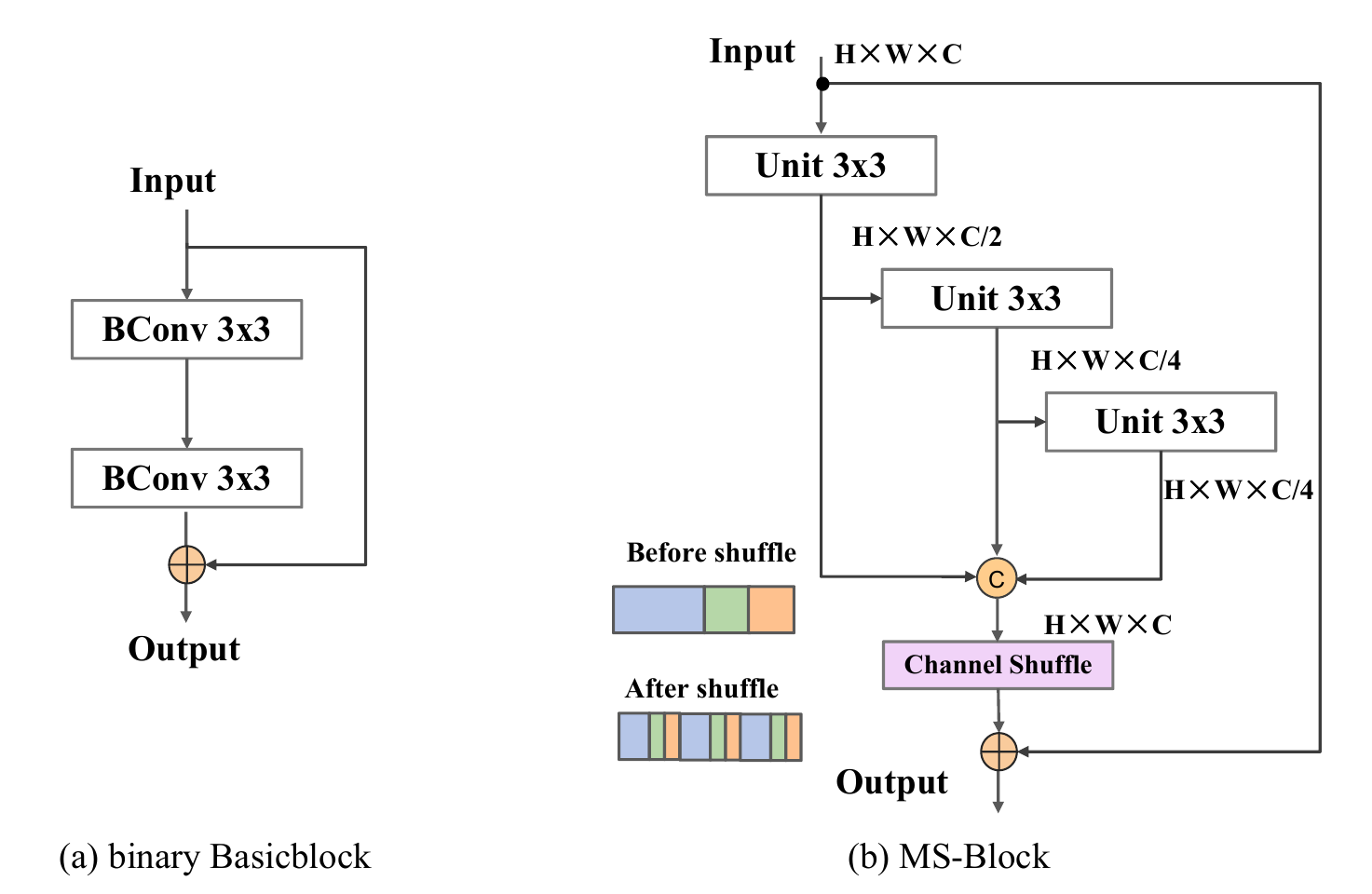}
\centering
\caption{\centering MS-Block design}
\label{FIG：4}
\end{figure}

Multi-scale block uses parallel convolution kernels of different sizes to extract information at different scales, and each branch has a different receptive field. The block allows the model to capture information at different scales and incorporate both local and global contextual information. By using multiple kernel sizes, the multi-scale block enhances the model's ability to extract features across different spatial resolutions, enabling the network to better understand and interpret complex patterns present in the input data. As a result, the multi-scale block contributes to the overall richness and robustness of the network's representations. However, the use of large-scale convolution kernels will increase the amount of parameters. To this end, 7×7 filter is decomposed to three 3×3 filters, and 5×5 filter is decomposed to two 3×3 filters. The input is divided into three branches, the first branch uses 3×3 filter, the second uses 5×5 filter, and the third uses 7×7 filter. This design effectively uses each convolution, and does not increase the complexity of the network while increasing the receptive field within the block. Each convolutional layer has a direct path connecting it to the output. We denote the features after 7×7 filter as $M_{A}=\left\{F_{0}^{A}, F_{1}^{A}, \ldots, F_{1 / 2 n-1}^{A}\right\}$ , $M_{B}=\left\{F_{0}^{B}, F_{1}^{B}, \ldots, F_{1 / 4 n-1}^{B}\right\}$ for 5×5, while $M_{C}=\left\{F_{0}^{C}, F_{1}^{C}, \ldots, F_{1 / 4 n-1}^{C}\right\}$ for 3×3, $n$ represents the number of channels.

However, the structure of the Multi-Scale block is still simple. Inspired by ReActNet\cite{Liu2020ReActNetTP}, we use more binarization-friendly Binary Unit module, the use of shortcut and PReLU can greatly improves the information carrying ability and representation ability of the network. In addition, we noticed that the direct concatenation of the outputs for the three convolutions will cause an unbalanced distribution of the output channels, where each output channel is only related to the output of the corresponding convolution. In order to solve this problem, we add a channel shuffle block after the concatenation to mix the channel information evenly. The features after shuffle block can be expressed as:  $\left\{F_{0}^{A}, F_{1}^{A}, F_{0}^{B}, F_{0}^{C},
\dots \allowbreak, F_{1 / 2 n-1}^{A}, F_{1 / 4 n-1}^{B}, F_{1 / 4 n-1}^{C}\right\}$ .

The proposed MS-Block is lightweight while efficient. Through MS-Block, global and local information at different levels and scales can be comprehensively considered, thereby improving the accuracy and robustness of the model. The block helps to model and predict the shape, structure and posture changes of  human bodies at different scales, making the network more accurate and reliable.

\section{Experiment}
\vspace{\baselineskip}
\begin{table*}[htbp]
\centering
\caption{Comparisons on COCO val set. \protect\#Params is the parameters of the pose estimation network. OPs represents the total number of operations, including floating-point operations FLOPs and binary operations BOPs, the calculation method is: OPS=BOPs/64+FLOPs}
\resizebox{\linewidth}{!}{ 
\begin{tabular}{@{}llccccccc@{}}
\toprule
\textbf{Model}  & \textbf{Backbone}  & \textbf{Bitwise(W/A)} & \textbf{Input Size} & \textbf{\#Params} & \textbf{OPs} & \textbf{AP}   & \textbf{AP50} & \textbf{AP75} \\
\midrule
Large Networks          &               &              &            &          &     &      &      &      \\
\midrule
SimpleBaseline\cite{Xiao2018SimpleBF}  & ResNet-50     & 32/32        & 256×192    & 34.0     & 8.9 & 70.4 & 88.6 & 78.3 \\
HRNet\cite{Sun2019DeepHR}            & HRNet-W32     & 32/32        & 256×192    & 28.5     & 7.1 & 73.4 & 89.5 & 80.7 \\
BHRNet\cite{He2022BinarizingBC}          & HigherHRNet   & 1/1          & 512×512    & -        & 7.9 & 60.6 & -    & -    \\
\midrule
Small Networks          &               &              &            &          &     &      &      &      \\
\midrule
Small HRNet\cite{Sun2019DeepHR}      & HRNet-W18     & 32/32        & 256×192    & 1.3      & 0.5 & 55.2 & 83.7 & 62.4 \\
MobileNetV2\cite{Sandler2018MobileNetV2IR}  & MobileNetV2   & 32/32        & 256×192    & 9.6      & 1.4 & 64.6 & 87.4 & 72.3 \\
ShuffleNetV2 1×\cite{Ma2018ShuffleNetVP} & ShuffleNetV2  & 32/32        & 256×192    & 7.6      & 1.2 & 59.9 & 85.4 & 66.3 \\
Lite-HRNet\cite{Yu2021LiteHRNetAL}       & Lite-HRNet-18 & 32/32        & 256×192    & 1.1      & 0.2 & 64.8 & 86.7 & 73.0 \\
                        & Lite-HRNet-30 & 32/32        & 256×192    & 1.8      & 0.3 & 67.2 & 88.0 & 75.0 \\
\textbf{BiHRNet (Ours)}          & \textbf{HRNet-W32}     & \textbf{1/1}          & \textbf{256×192}    & \textbf{9.9}      & 0.6 & \textbf{68.4} & \textbf{90.5} & \textbf{75.8} \\
\midrule
Small HRNet\cite{Sun2019DeepHR}      & HRNet-W18     & 32/32        & 384×288    & 1.3      & 1.2 & 56.0 & 83.8 & 63.0 \\
MobileNetV2 1×\cite{Sandler2018MobileNetV2IR}  & MobileNetV2   & 32/32        & 384×288    & 9.6      & 3.3 & 67.3 & 87.9 & 74.3 \\
ShuffleNetV2 1×\cite{Ma2018ShuffleNetVP} & ShuffleNetV2  & 32/32        & 384×288    & 7.6      & 2.8 & 63.6 & 86.5 & 70.5 \\
Lite-HRNet\cite{Yu2021LiteHRNetAL}       & Lite-HRNet-18 & 32/32        & 384×288    & 1.1      & 0.4 & 67.6 & 87.8 & 75.0 \\
                        & Lite-HRNet-30 & 32/32        & 384×288    & 1.8      & 0.7 & 70.4 & 88.7 & 77.7 \\
\textbf{BiHRNet (Ours)}          & \textbf{HRNet-W32}     & \textbf{1/1}          & \textbf{384×288}    & \textbf{9.9}      & 1.3 & \textbf{70.8} & \textbf{91.5} & \textbf{78.3} \\
\bottomrule
\end{tabular} }
\end{table*}

We use two datasets, COCO 2017\cite{Lin2014MicrosoftCC} and MPII\cite{Andriluka20142DHP}, to evaluate our method. Following the commonly used top-down framework\cite{Sun2019DeepHR}\cite{Luo2020EfficientHP}, our method estimates a heatmap of $K$ keypoints to represent the confidence of locations. We conduct comprehensive ablation experiments and report comparisons with other lightweight networks on two datasets.

\subsection{Setting}
\vspace{\baselineskip}
\subsubsection{Key layer settings}
Some layers has a greater impact on binary model performance compared with other layers, these layers require additional attention. For downsampling layers, image resolution will be reduced by half, the information loss in this process is irreversible. In order to avoid excessive loss of accuracy, this part uses real-valued calculations. The first layer of the network is real-valued to avoid huge information loss in the beginning. To avoid the influence of binarization and get accurate heatmap, the final layer also preserves real-valued weights and activations\cite{Martnez2020TrainingBN}\cite{Bethge2019BinaryDenseNetDA}\cite{Zhuang2018StructuredBN}. 

\vspace{\baselineskip}
\subsubsection{Training}
Our network is trained on two GeForce RTX 3090 GPUs, using the Adam optimizer to update all parameters. The initial learning rate is set to 1e-3, and is reduced to 1e-4 and 1e-5 at 170th and 200th epoch respectively. Training process stops at 210th epoch. We fix the height and width ratio of the human detection box to 4:3, and crop the detection box from the image, which is resized it to a fixed size: 256×192 or 384×288 on the COCO dataset, and 256×256 on the MPII dataset.

\vspace{\baselineskip}
\subsubsection{Testing}
We use a two-stage top-down approach, first use a person detector to detect the location of people in the image, and then detect keypoints in detection box. We use the same human detector as HRNet on the validation dataset. The position of each keypoint is obtained by shifting 1/4 pixel from the position of the highest response to the direction of the second highest response\cite{Sun2019DeepHR}\cite{Xiao2018SimpleBF}. 

\subsection{Microsoft COCO}
\vspace{\baselineskip}
\subsubsection{Dataset \protect\& Metrics}
The COCO dataset contains over 200K images and 250K person instances labeled with 17 keypoints. Our model is trained on the COCO train2017 dataset (including 57K images and 150K human instances). Model performance is evaluated on the val2017 dataset, which contain 5k and 20k images, respectively.
The evaluation metric of the COCO dataset is based on Object Keypoint Similarity (OKS):
\begin {equation}
\label{OKS}
\mathrm{OKS}=\frac{\sum_{i} \exp \left(-d_{i}^{2} / 2 s^{2} k_{i}^{2}\right) \delta\left(v_{i}>0\right)}{\sum_{i} \delta\left(v_{i}>0\right)}
\end{equation}
where  $d_i$ is the Euclidean distance between the detected keypoint and the ground truth,  $v_i$ is the sign of visibility the ground truth, s represents the scale factor of the target, and   is a constant that controls falloff of each keypoint. Based on OKS, we report AP(the mean of AP scores), AP50(AP at OKS=0.50), and AP75 as the experimental results.

\begin{figure*}
	\centering
        \includegraphics[width=\textwidth]{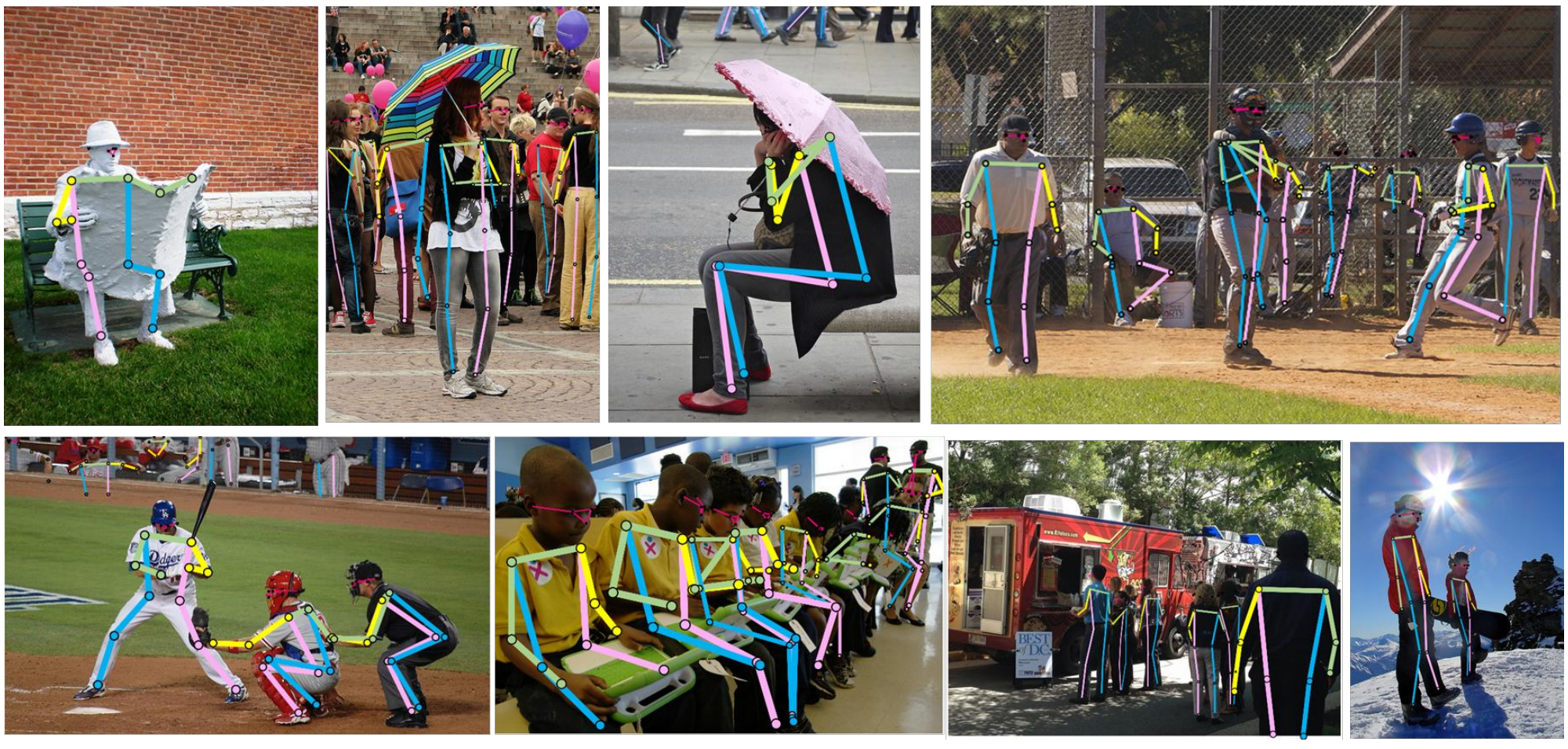}
	\caption{\centering  Example of human pose estimation on COCO}
	\label{FIG:5}
\end{figure*}

\begin{table*}[ht]
\centering
\caption{Comparisons on COCO test-dev set.}
\resizebox{\linewidth}{!}{ 
\begin{tabular}{@{}lllllllllll@{}}
\toprule
\textbf{Model}           & \textbf{Backbone}      & \textbf{Bitwise} & \textbf{Input Size} & \textbf{\#Params} & \textbf{OPs}  & \textbf{AP}   & \textbf{AP50} & \textbf{AP75} & \textbf{APM}  & \textbf{APL}  \\
\midrule
SimpleBaseline\cite{Xiao2018SimpleBF}  & ResNet-50     & 32/32   & 256×192    & 34.0     & 8.9  & 70.0 & 90.9 & 77.9 & 66.8 & 75.8 \\
HRNet\cite{Sun2019DeepHR}           & HRNet-W32     & 32/32   & 384×288    & 28.5     & 16.0 & 74.9 & 92.5 & 82.8 & 71.3 & 80.9 \\
Small HRNet\cite{Sun2019DeepHR}     & HRNet-W18     & 32/32   & 384×288    & 1.3      & 1.2  & 55.2 & 85.8 & 61.4 & 51.7 & 61.2 \\
MobileNetV2\cite{Sandler2018MobileNetV2IR} 1×  & MobileNetV2   & 32/32   & 384×288    & 9.8      & 3.3  & 66.8 & 90.0 & 74.0 & 62.6 & 73.3 \\
ShuffleNetV2\cite{Ma2018ShuffleNetVP} 1× & ShuffleNetV2  & 32/32   & 384×288    & 7.6      & 2.8  & 62.9 & 88.5 & 69.4 & 58.9 & 69.3 \\
Lite-HRNet\cite{Yu2021LiteHRNetAL}      & Lite-HRNet-18 & 32/32   & 384×288    & 1.1      & 0.4  & 66.9 & 89.4 & 74.4 & 64.0 & 72.2 \\
\textbf{BiHRNet (Ours)}  & \textbf{HRNet-W32}     & \textbf{1/1}     & \textbf{384×288}    & \textbf{9.9}      & \textbf{1.3}  & \textbf{68.3} & \textbf{90.2} & \textbf{76.1} & \textbf{65.2} & \textbf{73.9} \\
\bottomrule
\end{tabular}}
\end{table*}

\begin{table}[ht]
\caption{Comparison of the results on the MPII val dataset, OPs represents the total number of operations, including floating-point operations FLOPs and binary operations BOPs, the calculation method is: OPS=BOPs/64+FLOPs}
\resizebox{\linewidth}{!}{
\begin{tabular}{llccccc}
\toprule
\textbf{Method }  & \textbf{Params} & \textbf{FLOPs} & \textbf{BOPs} & \textbf{OPs} & \textbf{PCKh@0.5}  \\
\midrule
HRNet                   & 28.5        & 9.5   & -    & 9.5 & 90.3      \\
MobileNetV2 1×\cite{Sandler2018MobileNetV2IR}  & 9.6         & 1.9   & -    & 1.9 & 85.4      \\
MobileNetV3 1×\cite{Howard2019SearchingFM}  & 8.7         & 1.8   & -    & 1.8 & 84.3      \\
ShuffleNetV2 1×\cite{Ma2018ShuffleNetVP} & 7.6         & 1.7   & -    & 1.7 & 82.8      \\
Small HRNet\cite{Sun2019DeepHR}     & 1.3         & 0.7   & -    & 0.7 & 80.2      \\
Lite-HRNet-30\cite{Yu2021LiteHRNetAL}    & 1.8         & 0.4   & -    & 0.4 & 87.0      \\
Lite-HRNet-18\cite{Yu2021LiteHRNetAL}    & 1.1         & 0.2   & -    & 0.2 & 86.1      \\
\midrule
BinaryHPE1\cite{Bulat2017BinarizedCL}      & 6.0         & -     & -    & -   & 78.1      \\
BinaryHPE2\cite{Bulat2019MatrixAT}      & 6.0         & -     & -    & -   & 82.5      \\
\textbf{BiHRNet(Ours)}           & \textbf{9.8}         & \textbf{0.7}   & \textbf{1.5}  & \textbf{0.8} & \textbf{87.9}      \\
\bottomrule
\end{tabular}
}
\end{table}

\vspace{\baselineskip}
\subsubsection{Results on the COCO val2017 Set}
In order to facilitate a more intuitive comparison between our method and other approaches, we divided the comparison network into two categories: large networks and small networks. Additionally, to comprehensively evaluate the performance of our proposed algorithm across different scales, we conducted evaluations at two resolutions. We report the comparison of our method with other networks in Table 2. When the input is 256×192, our network obtains an AP score of 68.4, which is better than most real-valued lightweight networks, and achieves a good balance between efficiency and accuracy.  Our binary network has similar FLOPs to the full-precision network Small HRNet, and the AP exceeds it by 11 points. Compared with the real-valued HRNet, the parameter amount of the model is 35\% of the original, and the calculation consumption is 8\%, which shows the powerful ability to reduce calculational cost. Compared with the only work using binary network (using the bottom-up method) on COCO dataset, our network has higher accuracy with less than 1/10 of the computational cost. 
Compared with Lite-HRNet, our binary network achieves accuracy exceeding 1.2 points at small resolutions and 0.4 points at large resolutions. Our BiHRNet has similar FLOPs compared to other lightweight networks. It can be observed that our method does not have an advantage in the number of parameters.
This is because our method involves binarization, which results in a reduction in bit accuracy. However, it's important to note that the actual number of parameters in the network does not change. As a result, the number of parameters in our binarized network is greater than that of a carefully designed real-valued lightweight network. Nevertheless, the number of parameters for BiHRNet is 60\% lower than that of the original HRNet. Figure 6 shows qualitative pose estimation evaluations on COCO. It is observed that our binary model is still able to achieve reliable and robust pose estimation performace with various background clutters and different viewing conditions. 

\vspace{\baselineskip}
\subsubsection{Results on the COCO test-dev2017 Set}
Table3 reports the comparison results of our binary networks and other state-of-the-art real-valued methods. Our method achieves 68.3 AP score, which a better performance than the small networks. Although there seems to be no obvious advantage in the number of parameters, our network has lower OPs, which means our method has higher computational efficiency.

\begin{figure*}
	\centering
        \includegraphics[width=\textwidth]{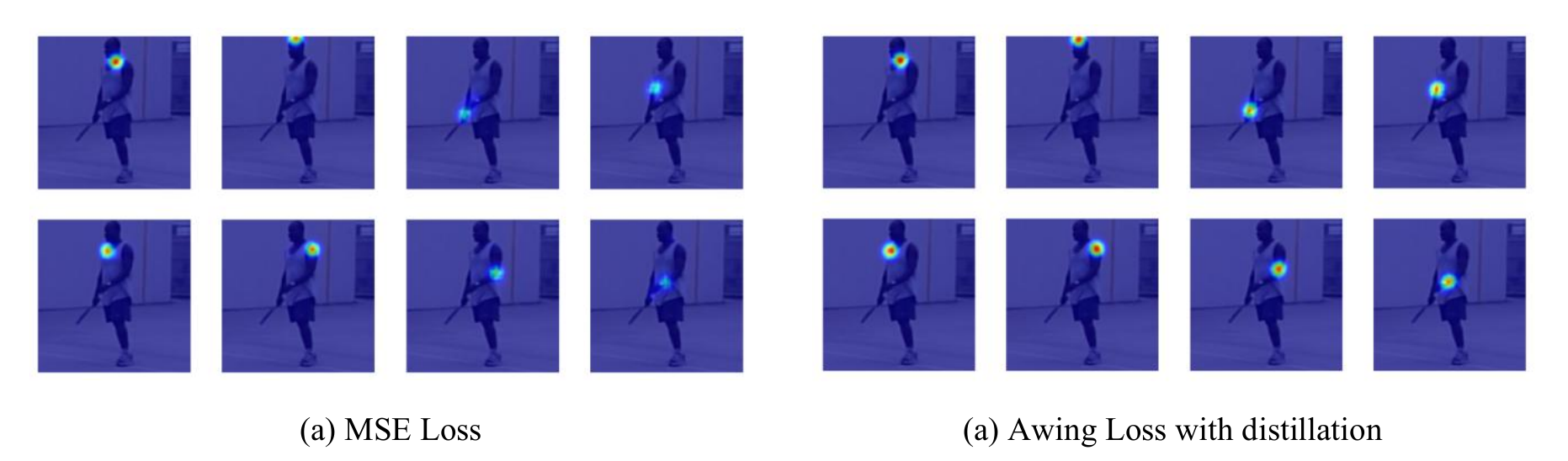}
	\caption{\centering  Heatmaps of different loss}
	\label{FIG:6}
\end{figure*}

\subsection{MPII Human Pose Estimation}
\vspace{\baselineskip}
\subsubsection{Dataset \protect\& Metrics}
The MPII dataset consists of images captured over a wide range of real-world activities, which annotated with full-body keypoints. The dataset contains approximate 25K images, including 40K subjects, of which 12K subjects constitute the test set, and the rest are used as the training set. The training strategy is the same as COCO dataset, except that the input size is cropped to 256×256 for fair comparison with other methods.

MPII uses PCKh as the standard evaluation metric, which means normalized distance is calculated using the person's head diameter as the scale factor. If the distance between the estimated position and the true value of a keypoint is within $\alpha 1$ pixels, it is regarded as a correct estimation, where $\alpha$ is a constant and $l$ is the head length. We report a score of PCKh@0.5 ( $\alpha$ = 0.5 ).

 \vspace{\baselineskip}
 \subsubsection{Results on the MPII val Set}
Table 4 shows the results of each model on MPII val. The resolution of the input images is 256×256. Our network is trained from scratch and obtains a PCKh score of 87.9. Compared with previous binarized pose estimation networks, our network outperforms these networks in accuracy. Compared with the real-valued HRNet, our method achieves a 66\% reduction parameters while requiring less than 1/10 of the original computational costs.

We also provide some results of other full-precision lightweight pose estimation networks for comparison. Compared with MobileNetV2, our method has an improvement of 2.5 in accuracy and has similar parameters, while computational cost is about half of that. Compared with other lightweight HRNet-based networks, although our method has a larger number of parameters, it has better accuracy performance, which proves the effectiveness of our method.

\subsection{Ablation Study}
\vspace{\baselineskip}

\begin{table*}[ht]
\caption{Structural Ablation Experiments on MPII}
\resizebox{\linewidth}{!}{
\begin{tabularx}{\linewidth}{@{}XXXXXX@{}}
\toprule
\multirow{2}{*}{\textbf{Arch}} & \multicolumn{3}{c}{\textbf{Setting}}     & \multirow{2}{*}{\textbf{PCKh@0.5}} & \multirow{2}{*}{\textbf{\#Params}} \\  \cline{2-4}
                      & \textbf{MS-Block} & \textbf{Pruning} & \textbf{IR BottleNeck} &                           &                           \\

\midrule
Binary HRNet          &         &         &               & 76.4                      & 28.5                      \\
BiHRNet               & \Checkmark     &         &               & 87.6                      & 11.5                      \\
BiHRNet               & \Checkmark       & \Checkmark       &               & 86.9                      & 9.8                       \\
BiHRNet(Ours)         & \Checkmark       & \Checkmark       & \Checkmark             & 87.5                      & 9.9     \\
\bottomrule
\end{tabularx} }
\end{table*}

To verify the effectiveness of the proposed structure and loss function, we conduct sufficient ablation experiments on the MPII dataset. These experiments demonstrate the effectiveness in reducing information loss in binary networks.

\vspace{\baselineskip}
\subsubsection{The effectiveness of structure}
After using the proposed MS-Block, the accuracy of the binary pose estimation network increases 11.2, and the parameter amount of the network is 40\% of the real-valued one, which shows that the designed multi-scale structure is high-performance and lightweight. The results are shown in Table 5.
To make the network simpler, we pruned the network following LitePose\cite{Wang2022LitePE}. After this adjustment, the accuracy is reduced, but the number of parameters is further reduced. On the basis of pruning, we replaced the structure of the first stage with the designed information reconstruction Bottleneck, and the accuracy increased by 0.6 when the number of parameters was almost unchanged, which shows that the structure we proposed is better than the initial stage of the network. The information is well preserved, which proves the effectiveness of the module.

\begin{table}[ht]
\caption{ Loss function ablation experiments on MPII val}
\begin{tabular}{lll}
\toprule
\textbf{Loss Function} & \textbf{PCKh@0.5} & \textbf{PCKh@0.1} \\
\midrule
Mse           & 86.3     & 31.0     \\
Awing         & 86.9     & 32.5     \\
Mse+KD        & 86.9     & 31.9     \\
Awing+KD      & 87.4     & 32.2    \\
\bottomrule
\end{tabular}
\end{table}

\vspace{\baselineskip}
\subsubsection{The effectiveness of loss function}
We observed that MSE loss will drop too low in later period of training. At this time, the distillation loss is an order of magnitude larger than the pose loss. For  $\alpha$ is 0.5, the overall loss will be approximately half of the KL loss. As a result, the optimization goal changes to learn the output of the real-valued teacher network, receiving few feedback from ground truth. The Awing loss has achieved better results in distillation training. It maintains the same magnitude as the KL loss, and can effectively learn the distribution information from the output label of the real-valued network.

The results of loss ablation experiments are shown in Table 6. The PCKh using Awing loss alone is 0.6 higher than MSE, which shows that Awing loss is more suitable for binary pose estimation network training. After using knowledge distillation, the accuracy of both loss functions has been improved. Among them, MSE has increased by 0.6, while Awing has increased by 0.5. Awing loss combined with knowledge distillation has achieved the best performance. As shown in Figure 6, the proposed loss function allows the binarized pose estimation network to obtain better heatmap responses.

\section{Conclusion}
We propose BiHRNet, a binary pose estimation network. We start from simply binarizing HRNet and optimize it step by step. We use knowledge distillation to make the output heatmaps of binary network closer to its real-valued counterparts. In addition, we design a loss function that is more suitable for binary HPE tasks. In order to reduce the information loss caused by network binarization, we focus on improving the basic modules that constitute the network: the bottleneck block and the basicblock. To reduce the information gap of the network at the initial stage, we design a binary bottleneck block for information reconstruction. To enhance the expressive power of the network while reducing the computational cost of the network, we design MS-Block. Our proposed network inherits the advantages of high-resolution networks and binary networks. Extensive experiments show that BiHRNet is effective and efficient.

\section{Acknowledgment}
This work was supported partly by the National Natural Science Foundation of China (Grant No. 62173045, 62273054), partly by the Fundamental Research Funds for the Central Universities (Grant No. 2020XD-A04-3), and the Natural Science Foundation of Hainan Province (Grant No. 622RC675).

\bibliographystyle{unsrt} 
\bibliography{cas-dc-template}

\bio{people1.pdf}
{\bf Zhicheng Zhang}\
received the B.S. degree in information engineering, M.S. degree in pattern recognition, and Ph.D. degree in control theory and engineering from Jilin University, Changchun, China, in 2005, 2007 and 2011 respectively. 
From 2011 to 2018, he was a teacher with the Control Science and Engineering Department, Jilin University, Changchun, China, where he had been an Associate Professor since 2015. Since 2018, he has been an Associate Professor with the School of Artificial Intelligence, Beijing University of Posts and Telecommunications, Beijing, China. His research interests include blockchain technique, machine learning, and swarm intelligence.
\endbio

\bio{people2.pdf}
{\bf Xueyao Sun}\
 received the bachelor's degree in Beijing University of Posts and Telecommunications, Beijing, China in 2021, and is currently pursuing the master's degree in Beijing University of Posts and Telecommunications, Beijing, China. Her current research interests include machine learning and computer vision. 
\endbio
\vspace{1\baselineskip}
\vspace{1\baselineskip}
\vspace{1\baselineskip}

\bio{people3.pdf}
{\bf Yonghao Dang}\
received the doctor's degree in Beijing University of Posts and Telecommunications and the bachelor's degree in computer science and technology from the University of Jinan, Jinan, China, in 2018. He is currently a postdoctoral researcher at Beijing University of Posts and Telecommunications, Beijing, China. His research interests include computer vision, machine learning and image processing, and deep learning 
\endbio

\bio{people4.pdf}
{\bf Jianqin Yin}\
received the doctor's degree from Shandong University, Jinan, China, in 2013. She currently is a Professor with the School of Artificial Intelligence, Beijing University of Posts and Telecommunications, Beijing, China. Her research interests include service robot, pattern recognition, machine learning and image processing.
\endbio

\end{sloppypar}
\end{document}